\documentclass[a4paper,12pt]{article}
\usepackage[english]{babel}
\usepackage[affil-it]{authblk}

\usepackage[letterpaper,top=2cm,bottom=2cm,left=3cm,right=3cm,marginparwidth=1.75cm]{geometry}

\usepackage{amsmath}
\usepackage{amssymb}
\usepackage{cite}  
\usepackage{graphicx}
\usepackage[usenames,dvipsnames]{color}
\usepackage[colorlinks=true, allcolors=blue]{hyperref}
\usepackage[justification=centering, font=footnotesize]{caption}
\usepackage{booktabs}
\usepackage{multirow}
\usepackage{listings}

\definecolor{comment}{rgb}{0, 0.6, 0}
\definecolor{keyword}{rgb}{0.96, 0.52, 0.02}
\definecolor{string}{rgb}{0.12, 0.4, 0.87}
\definecolor{background}{rgb}{0.95,0.95,0.92}
\lstdefinestyle{mypython}{
  language=Python,
  backgroundcolor=\color{background},   
  commentstyle=\color{comment},
  keywordstyle=\color{keyword},
  numberstyle=\tiny\color{magenta},
  stringstyle=\color{string},
  basicstyle=\ttfamily\footnotesize,
  breakatwhitespace=false,         
  breaklines=true,                 
  captionpos=b,
  keepspaces=true,
  numbers=left,
  numbersep=5pt,
  showspaces=false,
  showstringspaces=false,
  showtabs=false,
  tabsize=2
}
\usepackage{academicons}
\usepackage{xcolor}

\newcommand{\myfig}[1]{{\color{black}Figure~\ref{#1}}}

\newcommand{\mycite}[1]{{\color{black}\cite{#1}}}
\newcommand{\mycode}[1]{{\color{black}Listing~\ref{#1}}}
\newcommand{\inlinecode}[1]{{\ttfamily\footnotesize{#1}}}
\newcommand{\auth}[1]{{\normalsize#1}}
\newcommand{\inst}[1]{{\small#1}}

\title{Functional Programming Paradigm of Python for Scientific Computation Pipeline Integration}

\author[1]{
  \auth{Chen Zhang}
  \thanks{E-Mail: \texttt{chen.zhang\_06sept@foxmail.com};}
}
\author[1]{\auth{Lecheng Jia}}
\author[2]{\auth{Wei Zhang}}
\author[3]{\auth{Ning Wen}}
\affil[1]{\inst{Real-time Technology Laboratory, Shenzhen United Imaging Research Institute of Innovative Medical Equipment}}
\affil[2]{\inst{Bussiness Division of Radiotherapy, United Imaging Healthcare Inc.}}
\affil[3]{\inst{Institute for Medical Imaging Technology, Shanghai Jiao Tong University School of Medicine, United Imaging Healthcare}}

\date{}

\begin{document}
\maketitle

\begin{abstract}
As the preferred programming language for artificial intelligence (AI), Python has gained immense popularity due to its versatile syntax, flexibility of multiple programming paradigms, as well as the robust community support. This popularity becomes even further in the realm of scientific computation, machine learning, or AI algorithms where there is a great demand of data processing. Nevertheless, modern data processing unveil increasingly tendency of interdisciplinarity, which frequently involves in importing different technical approaches. An unified data control, is therefore in urgent, to gain performance of system integration among varying libraries. This integration is of the profound significance of accelerating prototype verification, optimizing algorithm performance, and minimizing maintenance cost.\par

Most researchers still have to make trade-offs between the ease of prototype implementation, and completeness of data engineering. For quite a few labs, using Python as a glue language, utilizing its advantages of abundance of third-party libraries, are preferable to accelerate in implementing prototypical algorithm dealing with complicated tasks. However, the lack of standardized data management, controlling, possibly pins their achievements on experimental stage, instead of contributing for practical usage. To tackle this issue, we proposed a frame of functional programming tailored for data processing pipelines. Utilizing this framework, we hope to achieve the following objectives:\par

\begin{itemize}
\item[1)]
Facilitate data processing and analytical tasks by leveraging a plethora of meta toolkit of scientific computation. By providing a unified interface, the framework enables seamlessly integrating existing functions, or even any callable object in Python, in highly engineered form. These features accelerate the construction for standard functional components, used both in researches, and data engineering.

\item[2)]
Establish the units of data flow controlling with concise syntax and unification application programming interface (API). The predefined operators of these units ensure consistency without conflict in component connection, therefore contribute to the creation of more systematic and complicated data processing pipelines, through existing functional units.

\item[3)]
Increasing the readability and maintainability of logical units via massively utilizing the closure functions as decorators. This design is featured as decoupling the definition for operations on data, from the assignment in arguments that affects these operations. Once a system becomes enormous, both the data operations and the arguments are tough to be managed in the design of object-oriented programming (OOP), while the one in case of this frame results seldom in mess.

\item[4)]
Proposing a scalable solution for data science and engineering, ensuring reliable performance across full ranges, from exploratory research to agile development and deployment. The extensibility of this framework guarantees its adaptation to the evolving requirements of data-driven projects, as well as to the increasing complexity of a data processing system.
\end{itemize}
\end{abstract}


\section*{Introduction}

Python gains increasing popularity since its compact but efficient design in recent years. As an interpreted programming language, Python is of the characteristics  of concise  syntax, flexibility on  multiple programming paradigms, cross-platforms, etc.\mycite{munawar2022impact, beazley2009python, summerfield2009programming}. Its software ecosystem gets enriched, which benefits from the  contributions of soaring researchers and developers studied in various fields. As with the booming development of AI techniques generally, Python has become a de facto development standard for scientific computation and AI algorithms due to not only some excel high performance libraries such as numpy, scipy, and tensorly\mycite{harris2020array, virtanen2020scipy, jean2019tensorly}, but also high compatibility for integrating other programming languages.\par

Nonetheless, as the expense of exceeding flexibility and the richness of software ecosystem, it gets also challenging in integration for complicated project via Python, particularly in scientific computation used in interdisciplinary application\mycite{oliphant2007python, perez2010python, turner2018applied, oxvig2016storing}. Such as, most of data manipulations are generally in demands of importing third party libraries designed on basis of different specifications, which can frequently result in incompatibility problems raised by data types or such like. Or, even if there is no compatibility issue raised by the data, when using interfaces with the identical function sourced from different packages (e.g. Gaussian filters in Pillow and opencv-python), factors such as different argument design or predefinition can further lead to variation in the final export. Similar situations, are generally too cumbersome to troubleshoot at the function invocation level, but have a major impact on the systematic reliability of our computation and analysis.\par

Those difficulties become even tough in the condition of operations involved in introducing complex algorithms, heavy numerical computations, or the manipulations on the great scale of dataset\mycite{cielen2016introducing, rodriguez2016general}. The conventional OOP approaches can become cumbersome and inefficient in such circumstances, as it may not offer the necessary flexibility and scalability to handle the complexity of scientific computing, especially during the exploring stage when functions or scripts will need frequent modification for refactoring.\par

In order to solve the above-mentioned challenges, based on a large number of practices, we have summarized a set of flexible and effective generic infrastructure of Python function based on functional programming (FP), to ensure both the invoking legality and extensibility for heavy modification as well as for further integration. With this framework, we achieved significant improvements with respect to code readability, maintainability, and performance. The FP paradigm, with its emphasis on immutable data and pure functions without side effects\mycite{gabbrielli2023functional, sarcar2023functional, hudak1989conception, alic2016comparative}, provides a more natural approach to handling the intricacies of scientific computing. It is in advantages on creation for modular and highly decoupled code style, which is precisely the essential for achieving completeness of functions, and the further optimization on the software architecture as well. The reliability and consistency of functions, or components built on this frame can therefore be both guaranteed.\par

\section*{Background and Architecture}

\begin{figure}[htbp]
  \centering
  \includegraphics[scale=0.55]{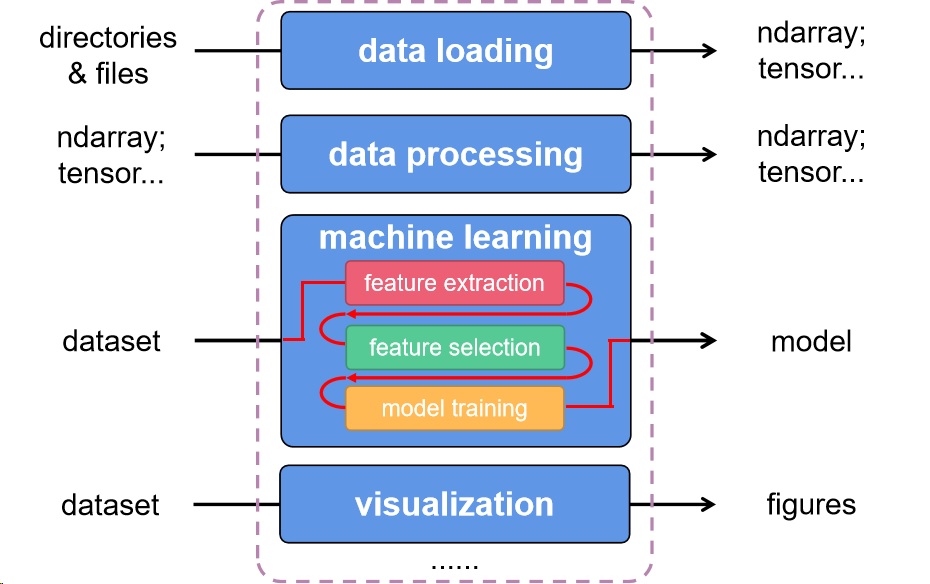}
  \caption{The essence of data manipulation: transforming data to represent it from one form to another.}
  \label{tag1}
\end{figure}

In the context of scientific computing, the most fundamental process is the transformation of data forms. In Python, this abstraction can be applied to a wide range of computing and analysis scenarios, including data pre- and post-processing, as well as machine learning model training (as illustrated in \myfig{tag1}). Despite differences in code or script organization, for identical functional units, they should at least include declarations of the types of data that can be accepted, the types of data that will be returned after manipulation, and so on. The Python programming language has already incorporated a similar feature since version 3.5, called typing hint\mycite{pep484}. This allows programmers to assign the suggested argument type when defining a function, although it acts more like an annotation than a static constraint.

\begin{figure}[htbp]
  \centering
  \includegraphics[scale=0.95]{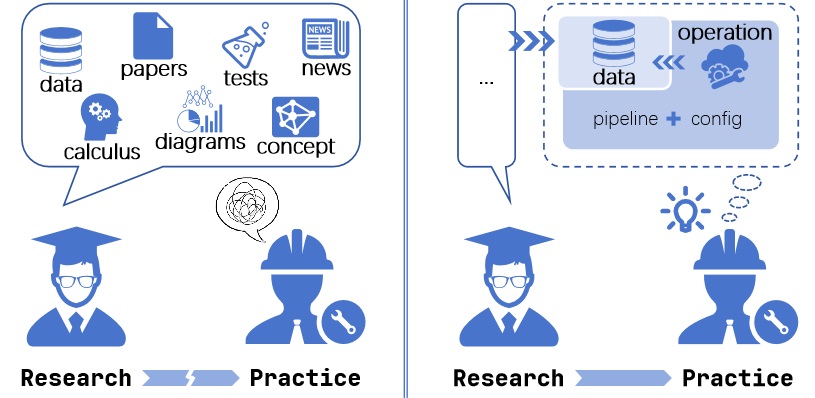}
  \caption{The technical divergence between data research and engineering}
  \label{tag2}
\end{figure}

The situation is further compounded in the domain of data exploration, where it is virtually impossible for an architecture to simultaneously satisfy the two contrasting requirements: the flexibility required by researchers in data investigation and the reliability required by engineers in implementation and deployment. The inherent technical divergence between these two populations can be illustrated in \myfig{tag2}. Researchers are generally more concerned with approaches and concepts than with their implementation, whereas data engineers are conversely more concerned with the implementation of concepts. This situation inevitably increases the communication costs in teamwork, as well as those in programming implementation and maintenance, which represents a significant obstacle to the transformation of scientific and technological achievements.\par

As for the majority of data analysts, Python is regarded as an essential tool to master due to its robust community support and plethora of exemplary third-party libraries. As Python is a dynamic language with a high degree of flexibility, it is important to implement a run-time check for passing in and out in order to avoid potential errors such as parameter type confusion or improper call. This can further reduce the potential risk for calling and integrating functions.  Based on the aforementioned causes, we propose a specific functional programming (FP) frame for Python that introduces forced constraints on simple functions, thereby ensuring logical correctness in more complex constructions.\par

\begin{figure}[htbp]
  \centering
  \includegraphics[scale=0.55]{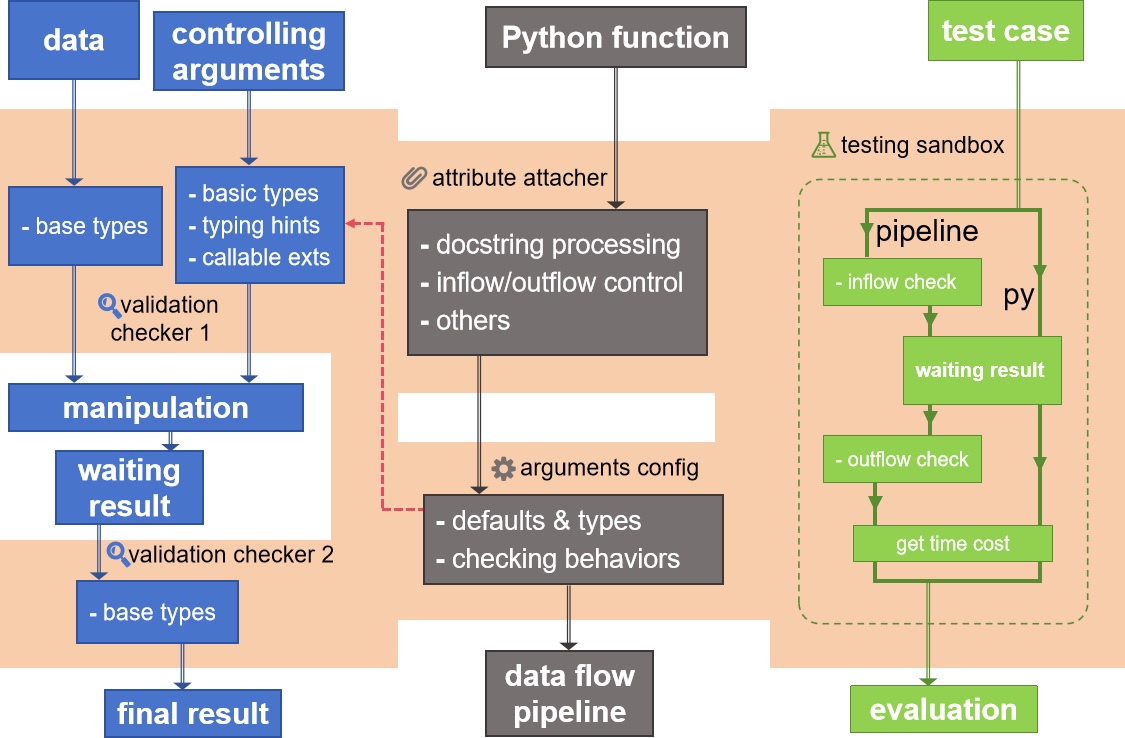}
  \caption{Illustration of the FP Python infrastructure used for the runtime checker (navy), function wrapper (gray), and tester (lime); the light salmon background is the framework's feature set.}
  \label{tag3}
\end{figure}

The illustration of this FP framework is as shown in \myfig{tag3}: the light salmon background represents the main scope of feature set of our infrastructure for Python functions, while blocks with different colours demonstrate the main function suites. In terms of type inspection (navy), the framework wraps the actual data manipulation inside, where data and corresponding arguments are validated before passing to the real execution, while the processed data is also checked again before returning. As a function or callable object wrapper (shown in gray), the frame first modifies the attributes of the decorated object, defines the data with the desired types in input and return, or even more strictly, configures the default values and their respective types of the function's associated arguments. The tester (shown in lime) is a development tool in function construction, used for executing case tests, logic validation, time-consuming evaluation, and so forth.\par

The design philosophy of this architecture is to logically isolate data manipulation into the data itself and the corresponding parameters for control, and then to technically separate the data and arguments from the body of the function definition. Its structure allows set operations on arguments of different decorated functions, which is crucial to support the construction of pipelines for more complicated structures. With regard to the manner of application, most of tools within this framework are closure functions employed as decorators in Python, therefore the utilisation of these features will not result in any additional learning costs for the majority of users. Furthermore, the necessity for extensive refactoring of the original project is also negated. The incorporation of these features facilitates the design and construction of data processing functions at all stages. The use of these wrappers enables function designers to conceal the inner details of the functions, while providing a set of standard argument configuration, akin to an API, in the function's pre-definition, which is then exposed to target users.\par

\section*{Function suite}

It is first necessary to explain the basic concept of Python's argument passing mechanism. The example function in \mycode{tag4} shows the four different types of parameters listed in order of inherent priority: positional argument(s), optional argument(s) with default value(s), var-positional, and var-keyword arguments. The prefix ``var'' means the uncertain length as a variable. In the context of Python syntax, it is not necessary to define all argument types simultaneously; however, once they have been defined, their relative order must be maintained in accordance with the aforementioned enumeration.\par

\begin{lstlisting}[language=Python, caption=Python arguments passing mechanism., label=tag4, firstline=1, lastline=9]
def function1(pos: int, opt: int = 1, *args, **kwargs):
    print(pos, opt, len(args), len(kwargs))
    return pos + opt + len(args) + len(kwargs)


function1(2)                 # 2, 1, 0, 0
function1(2, 3)              # 2, 3, 0, 0
function1(2, 3, 4, 5)        # 2, 3, 2, 0
function1(2, kw1=5, kw2=6)   # 2, 1, 0, 2
\end{lstlisting}

The call examples shown in the \mycode{tag3} section can assist us in comprehending the assignment behavior of arguments in the Python programming language. Python's syntax permits the utilization of arguments in two distinct manners: direct pass-in by values and declaration for their keywords during assignment. However, it is not possible to interchange the relative positions of these two assigning approaches (i.e., no-keyword calling prior to keyword one). In the absence of a keyword declaration, the interpreter attempts to fill the positional type first, followed by the optional type by overwriting. If there are still unmatched arguments, they are packed into the var positional type as a tuple. In contrast, arguments declared with keywords are processed as a dictionary at once.\par

In theory, any type of argument can be replaced in the same manner as the var-keyword. For instance, all types of arguments that appear in the example function in \mycode{tag3} can be encapsulated within a unified set of dictionary parameters, as illustrated in \mycode{tag5}.\par

\begin{lstlisting}[language=Python, caption=Equivalent wrapper using only var-keyword arguments., label=tag5, firstline=12, lastline=18]
def function2(**kw):
    return function1(
        kw.get('pos'),
        kw.get('opt', 1),
        *kw.get('args', ()),
        **kw.get('kwargs', {}),
    )
\end{lstlisting}

As our demonstration encompasses all potential forms of argument passing, the wrapper utilising solely the var-keyword is theoretically universal in its capacity to support any callable object within the Python programming language. The concept of the var-keyword function is of paramount importance within our architectural framework, which is designated the \emph{info function} (derived from the informatics project, as referenced in\mycite{informatics}). This designation will be reiterated below. The advantage of utilising the info function is that it enables the utilisation of features designed exclusively for var-keyword callable objects.\par

\subsection*{Attaching attributes}

With regard to the information function, it is typical for the keyword ``data'' to be reserved as the object to be processed in the current step, with the processed result being returned. In order to ensure that the accepted data is of the desired types, it is possible to simply prepend a decorator to the function with the appropriate typing hint declarations. To illustrate, when scripting a function for text manipulation, if it is desired to support processing for a string or a list composed of a set of strings, the corresponding constraint can be implemented in a concise manner as demonstrated in the code snippet labelled \mycode{tag6}. From the function call examples, it is evident that type checking is also performed on inner elements if an iterable complex structure is detected.\par

\begin{lstlisting}[language=Python, caption=Data inflow control via Python typing hint., label=tag6, firstline=20, lastline=37]
from info.me import FuncTools
from typing import Union, Any


@FuncTools.attach_attr(info_func=True,
                       entry_tp=Union[str, list[str]],
                       return_tp=Any)
def func1(**kw):
    _ = kw.get('data')
    ...
    return 0


func1(data='string0')               # valid, str
func1(data=['string1', 'string2'])  # valid, list[str]
func1(data=('string3', 'string4'))  # invalid, tuple[str, ...]
func1(data=50)                      # invalid, int
func1(data=['string5', 50])         # invalid, list[str | int]
\end{lstlisting}

The innovative aspect of the frame design is the logical decoupling of the argument validation steps from the main body of the function, which is achieved at a minimal cost. Without the attachment of the inflow control attribute (lines 5 to 7 in \mycode{tag6}, mainly 6), it is unsafe to assume the actual type it accepts in function entrance (line 9), therefore the check logic as shown in \mycode{tag7} must be embedded after where the data was acquired. The code snippet executed for similar purposes is technically referred to as defensive programming\mycite{BOULANGER2016125}. Despite its prevalence in modern software design, integrating this block into the function body sacrifices flexibility in dealing with varying requirements, while simultaneously reducing readability and maintainability of the code. To illustrate, in order to accommodate the case presented in line 16 of \mycode{tag6}, the conventional approach necessitates the incorporation of support for tuple containers synchronously, based on the original type check snippet. Conversely, for the info function, a straightforward replacement is sufficient. The value of \inlinecode{entry\_tp}, which is defined as \inlinecode{Union[str, list[str]]}, should be replaced with the value \inlinecode{Union[str, list[str], tuple[str, ...]]}. The example presented here serves to illustrate the control of data input. In contrast, for the control of data output, the similar type syntax applies to the argument \inlinecode{return\_tp}.\par

\begin{lstlisting}[language=Python, caption=Defensive programming snippet for argument checking., label=tag7, firstline=39, lastline=45]
...
if not isinstance(_, (str, list)):
    raise ValueError('...')
elif isinstance(_, list):
    if not all([isinstance(_v, str) for _v in _]):
        raise ValueError('...')
...
\end{lstlisting}

As a practice of functional programming, this architectural approach can not only serve software security well, but also contribute to the design of the software architecture itself. For many mature Python packages, the majority of their API function bodies consist of considerably long scripts, which include not only essential defensive programming snippets for all passed parameters, but also corresponding documentation and invocation examples. However, as for the info function, it is architecturally supported to introduce and then integrate those necessary attributes using outer objects. As illustrated by the pseudo-code in \mycode{tag8}, the documentation of \inlinecode{main\_func} can be fulfilled by a carrier function, \inlinecode{doc\_func}, defined elsewhere, rather than by direct insertion into the main body of \inlinecode{main\_func}.\par

\begin{lstlisting}[language=Python, caption=Attaching documentation through Python function carrier., label=tag8, firstline=48, lastline=58]
def doc_func(**kw):
    """
    documentation begin
    ... rst doc for main_func ...
    documentation end
    """
    ...

@FuncTools.attach_attr(docstring=doc_func)
def main_func(**kw):
    ...
\end{lstlisting}

Furthermore, additional attributes can be attached to the target function, such as scripting author(s), maintainer(s) or tester(s), during practical development as needed. This can be achieved with relative ease (see \href{https://informatics.readthedocs.io/en/latest/interface/api\_frame.html#customize-function-by-decorator}{examples}). It can be observed that the frame is characterised by the flexible use of Python functions, which fully increases system reliability without sacrificing its scalability. Consequently, this leads to further rigor in software construction.\par

\subsection*{Argument configuration}

The parameters of the info function are not solely comprised of the data to be processed; they also encompass the controlling arguments that can influence the processing steps. In general, the cases of controlling arguments are more intricate than that of data itself. \mycode{tag9} reveals three typical modes of argument configuration using our functional frame. Firstly, built-in types, such as the \inlinecode{ndarray} type for \inlinecode{data}, can be used as constraints to restrict arguments. Secondly, typing hints, such as a tuple composed of float and integer for \inlinecode{arg1}, can be employed. Thirdly, callable objects, such as a coefficient between 0 and 1 for \inlinecode{arg2}, can be used as constraints to restrict arguments.\par

\begin{lstlisting}[language=Python, caption=Three typical approaches for argument configuration., label=tag9, firstline=61, lastline=69]
from info.me import FuncTools, T, Null
from numpy import ndarray


@FuncTools.params_setting(data=T[Null: ndarray],
                          arg1=T[(.3, 60): tuple[float, int]],
                          arg2=T[0.7: (lambda x: 0 < x < 1)])
def func2(**kw):
    ...
\end{lstlisting}

The last type represents a revolutionary advancement in the field of type checking workflows, particularly in the context of scientific computation. The mathematical implementation in data engineering is more rigorous than that of software engineering. To illustrate, consider matrix computation. On occasion, users may wish to impose certain constraints (such as symmetric, positive, or semi-positive definite) on the matrix to be acquired. Fortunately, these constraints can be easily defined by means of lambda functions. Once the constraint functions have been prepared, the target function can accept matrices with specific properties, using the frame with the form shown in \mycode{tag10}. Furthermore, the lambda functions are dumped as values (\inlinecode{\_sym}, \inlinecode{\_pos}, and \inlinecode{\_semi\_pos}) in script, allowing them to be reused in any context where necessary, in the form of common Python functions.\par

\begin{lstlisting}[language=Python, caption=Applying matrix constraints on decorated function., label=tag10, firstline=72, lastline=86]
import numpy as np


_sym = (lambda x: True if (x.ndim == 2 and
                           x.shape[0] == x.shape[1] and
                           np.allclose(x, x.T)) else False)
_pos = (lambda x: np.all(np.linalg.eigvals(x) > 0))
_semi_pos = (lambda x: np.all(np.linalg.eigvals(x) >= 0))


@FuncTools.params_setting(mat1=T[Null: _sym],
                          mat2=T[Null: _pos],
                          mat3=T[Null: _semi_pos])
def func3(**kw):
    ...
\end{lstlisting}

This design feature has the potential to enhance the reusability of constraint functions while also improving the readability of decorated functions. Moreover, it provides an alternative solution for Python functions that can accommodate future derived types. The formal name for supporting those derivatives is terminologically called ``duck typing''\mycite{ertl2012methods}. This concept is increasingly supported by features of modern programming languages, including \inlinecode{concept} in \inlinecode{C++20} and \inlinecode{GenericAlias} in Python\mycite{pep484, sutton2012design, pep585}. To illustrate, consider the concept of a matrix. While its mathematical definition is relatively straightforward, the concrete implementations in Python remain challenging to enumerate. However, in our case, we can conveniently fulfill this requirement with an anonymous function that checks whether the dimension is 2 after type conversion into a numpy array. This approach allows for the legal passing of matrices and 2-dimensional arrays of numpy, sparse arrays or matrices of scipy, and any other nested iterable containers that are consistent with the mathematical concept of a matrix.\par

\subsection*{Testing frame}

In addition to type control for input data and corresponding arguments, our infrastructure also provides a wealth of accessory tools, such as \inlinecode{default\_param} for dynamic argument assignment, or \inlinecode{exception\_logger} for capturing exceptional cases during execution, which support the serial requirements of agile development. In this context, the sandbox for execution case testing can drastically simplify script editing and organisation, since its feature of testing while scripting via function decorator. The testing framework can be applied not only to the info functions, but also to common Python functions in any form of argument declaration. For further details, please refer to the \href{https://informatics.readthedocs.io/en/latest/interface/api\_frame.html#decorator-to-do-function-testing}{application examples} provided by the API documentation.\par

This feature is beneficial for software architects working with Python-based tool chains. Typically, a given Python module can be decomposed into a number of callable components, with developers appending the test cases to the end of their script. As the number of necessary components increases, the length of the script under editing also increases, making it inefficient and potentially risky to repeatedly roll up and down on the script for each necessary test. In addition to the features of in-place testing on callable components, this frame also records the time consumed during the entire call cycle. Consequently, this infrastructure can not only provide convenience of validation for each component in the module, but also simultaneously evaluate basic performance on each component. This can be the reliable evidence for hot-path analysis or other possible optimisations in the future.\par

The testing frame can be employed at various stages of the architectural design and module construction processes. Additionally, it can be utilized in unit testing, if necessary. To illustrate, the \href{https://github.com/CubicZebra/informatics/tree/main/test}{test directory} of the informatics project contains numerous test examples for the info functions, which are executed via a pipeline of unit tests with the test frame embedded.\par

\section*{Analysis and discussion}

In this section, we will further elucidate the reasons why our FP infrastructure and its application can contribute to pipeline integration. We will examine the features of the functional programming paradigm, the performance loss after function decoration, and the convenience brought by the pipelining code style.\par

\subsection*{Flexibility of FP}

The concept of a programming paradigm is not immediately intuitive. In essence, OOP represents a conceptual abstraction of entities with a noun attribute. This approach is analogous to the scientific classification of organisms, where the context domain, kingdom, phylum, class, order, family, and genus are used to describe creatures. OOP is a rigorous and comprehensive approach, but it lacks flexibility in dealing with open sets, such as the description of a completely new species that cannot be categorized by this system.\par

It is unfortunate that openness is precisely what matters in data exploration. The process of reading data from files is referred to as \emph{data loading}, the manipulation of data is defined as \emph{preprocessing}, the extraction of information from data is termed \emph{feature engineering}, and the training of data to identify underlying patterns is referred to as \emph{model training}. All of these operations have the verb attribute rather than that of noun. These behaviours exist objectively, but cannot be abstracted by the OOP paradigm without the input of concrete data for processing.\par

The appeal of FP lies in its capacity to define hitherto unknown behaviours, which can result in functions with high expressiveness and flexibility. For further illustration, consider the example of the \href{https://informatics.readthedocs.io/en/latest/tutorial/tut\_inflow.html#meta-data-loader}{generic data loader} as presented in the informatics project tutorials. This is an integration pipe. The high-order functions \inlinecode{search\_from\_root} and \inlinecode{generic\_filter} can be employed as an implementation of the generic data loader, despite the fact that the data to be loaded is unknown.\par

\subsection*{Performance loss}

\begin{figure}[htbp]
  \centering
  \includegraphics[scale=0.45]{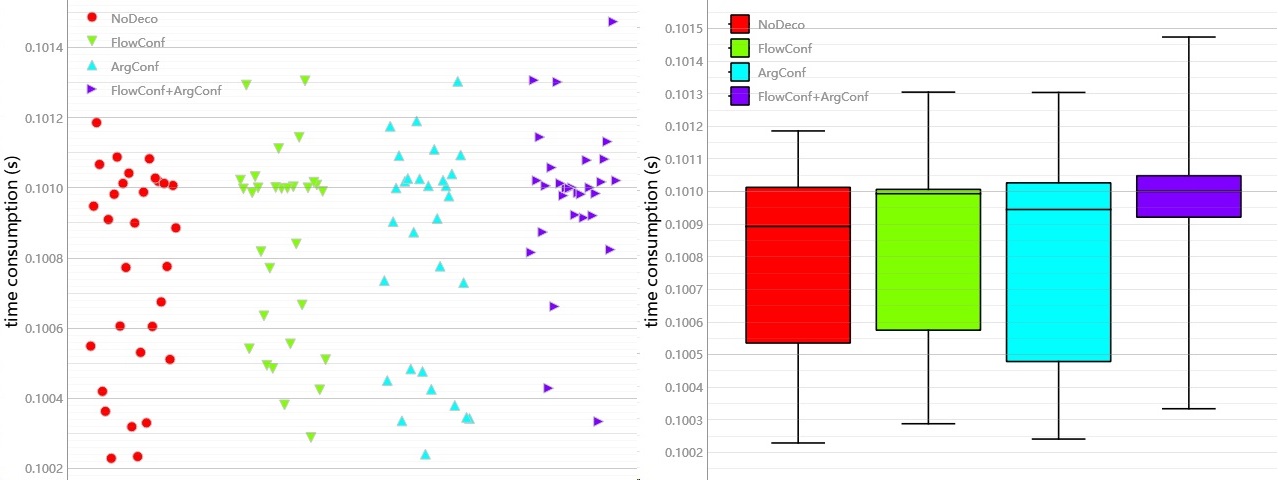}
  \caption{Beeswarm and box plots of benchmark tests on invocation time consumption, for an identical simple functions. \textbf{NoDeco}: without decorator; \textbf{FlowConf}: decorated by inflow/outflow control; \textbf{ArgConf}: decorated by argument configuration; and \textbf{FlowConf+ArgConf}: with both decorators.}
  \label{tag11}
\end{figure}

Given that the majority of the features in our programming framework applied to Python function are achieved through the use of decorators, it is reasonable to conduct a comprehensive evaluation of the performance loss when applying decorators. To this end, we have designed a benchmark test using a simple Python function with the behavior of hanging up for 0.1 seconds to mimic actual data processing, followed by a return. Two types of decorators were employed in this study: inflow/outflow control and argument configuration. These were applied separately or simultaneously to the original function, resulting in the generation of new callable objects. Each object was then subjected to 30 trials, after which their average time consumption was compared. The corresponding results were summarized in \myfig{tag11}.\par

The majority of the call instances are situated in the vicinity of the 0.1010 seconds position, as indicated by the scatter points in the beeswarm plot. Despite the varying degrees of numerical increase in function call time subsequent to decoration, their values are relatively close, with only minor discrepancies. Therefore, it is justifiable to accept a slight reduction in performance in order to guarantee the security and dependability of the input data and the parameters passed.\par

In particular, the core idea of the framework design is to prioritize data mapping over legal invocation for a specific computational step (although it can be utilized in this manner). It is also the distinction in design concept from the element-wise based high-order Python functions like map-reduce, as well as the essence of the FP paradigm. As illustrated in \myfig{tag1}, the majority of data manipulation within a computer system can be described as a process of mapping data from one existential form to another. This concept forms the basis of our proposed infrastructure, which can be viewed as an invocation protocol for data and arguments in data mapping. In its application, the user level may be more concerned with the content within the protocol itself, rather than the concrete logic of the function body. However, both aspects should be fully considered at the developer level.\par

\subsection*{Code style of pipeline}

Thus far, we have expounded at length on the advantages of our infrastructure for secure invocation and good code organization. However, there is a deeper significance that extends beyond these benefits. By employing a unified protocol designed for both data and argument passing, we can ensure that the entries and exits of functions will have clear definitions. Consequently, the decorated functions will be equipped with the ability to communicate with each other in the absence of any practical data. This feature is particularly advantageous for the design of computer systems and the integration of scientific computation pipelines, particularly in the absence of data.\par

The framework allows for the highly reusable integration of either the info function or the pipeline derived from info functions. This can be illustrated by considering a concrete computer vision (CV) task. In natural image processing, a pipeline including cropping, denoising and resampling is required. As these atomic operations are already implemented in the informatics project, they can be easily connected using the info functions wrapped by the inline code \inlinecode{Unit}. The code structure would be represented by the \mycode{tag12} (lines 4 to 5), where the ellipses within the list indicate the positions of the corresponding info functions. Similarly, if one were to compare different edge detection methods following this preprocessing operation, one could reuse this \inlinecode{processing} pipe in the definition of the experimental pipeline (lines 6 to 7 in \mycode{tag12}).\par

\begin{lstlisting}[language=Python, caption=Integrating scientific computation flows., label=tag12, firstline=89, lastline=95]
from info.me import Unit


crop, denoise, resample = [Unit(mappings=[_]) for _ in [...]]
processing = crop >> denoise >> resample
prewitt, canny, log = [Unit(mappings=[_]) for _ in [...]]
compare_experiment = processing >> (prewitt | canny | log)
\end{lstlisting}

\begin{figure}[htbp]
  \centering
  \includegraphics[scale=0.58]{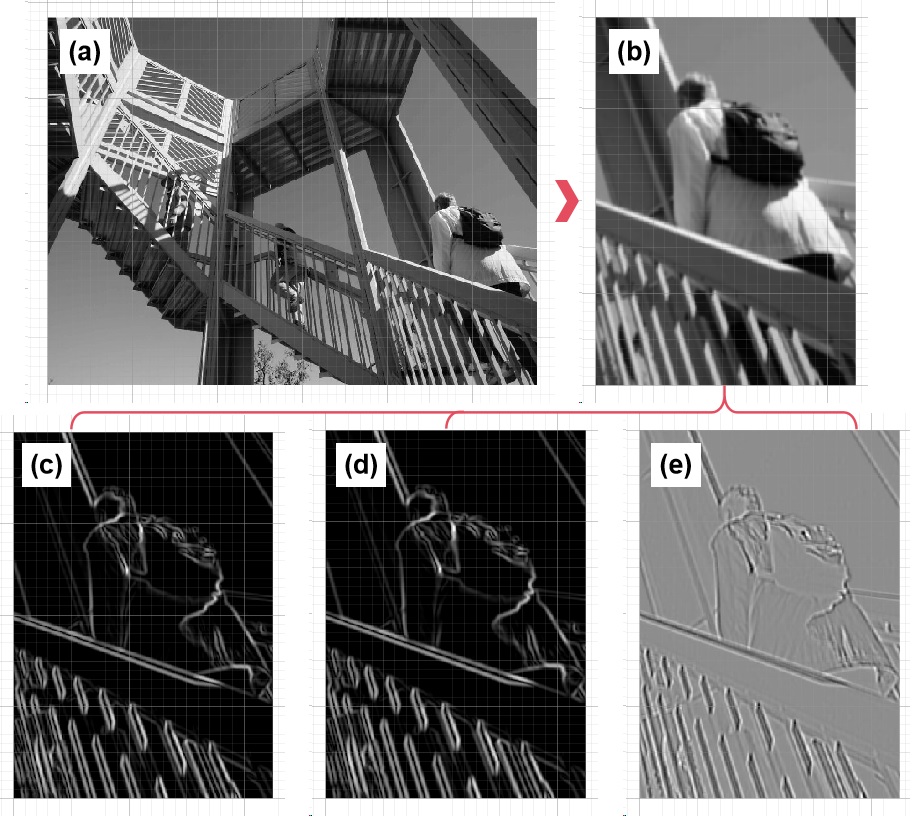}
  \caption{Demonstration of experimental pipe applied to natural image processing. (a) original 8-bit grayscale image; (b) preprocessed image after cropping, denoising, and resampling; and final images filtered by (c) Prewitt; (d) Canny; and (e) Laplacian of Gaussian.}
  \label{tag13}
\end{figure}

To demonstrate the functionality of the pre-processing pipeline, an 8-bit greyscale image was used as a test case. The image was pre-processed using a set of parameter configurations, and the intermediate image and edge-enhanced results after convolving the image with different kernels were generated. These results are shown in \myfig{tag13} (b) to (e), respectively. The example image was derived from a demonstration instance in the scipy package. The wrapped units and integrated pipeline can be perceived as a series of atomic operations of mappings on data.\par

Pipeline integration based on atomic operations can be conceptualised as the design of a data assembly line, with basic components analogous to the material flow in this data assembly line. In light of the growing importance of data science and artificial intelligence, researchers are increasingly reliant on interdisciplinary analysis of data in different modes. The establishment of a unifying criterion for programming paradigms is of profound importance now and in the future, as it ensures that data can flow seamlessly throughout prototypical tools designed in different disciplines, analysis frameworks and corresponding libraries.\par

\section*{Conclusion}

This paper presents a novel FP paradigm based on the Python architecture, designed for the integration of pipelines of different data mapping operations. In particular, it is intended for the integration of scientific computation flows, which are commonly involved in the applications of computational tools and the validation of prototypical ideas.\par

As a fundamental concept within the framework, we proposed a specific form of function derived from the generic Python function, which has been demonstrated to be of theoretical equivalence to that of the general one. With a suite designed for that specific Python function, we logically separate a Python function into factors such as the main execution body, the target data, the corresponding argument configuration, the attributes, and so forth. The framework employs the features of the functional programming paradigm and high-order functions to provide a multitude of meta-implementations with considerable flexibility, thereby affording analysts and researchers a wide range of options for addressing problems. For engineers and maintainers, the framework also offers advantageous characteristics, including forced type checking, inflow/outflow control, and the prevention of illegal invocations within functions. This facilitates the tracking and subsequent resolution of issues.\par

Furthermore, it supports a distinctive connection syntax when the atomic data operations are wrapped as data processing units. With our infrastructure for Python functions and the informatics project, we hope researchers can make full use of these sophisticated tools to achieve their creative and flexible goals, which will then lead to the valuable transformation of technological achievements.\par

\section*{Acknowledgements}

Our sincerest gratitude to the United Imaging Healthcare Group, the Shanghai Jiao Tong University School of Medicine, the Human Resources and Social Security Administration of Shenzhen Municipality, and the Human Resources Bureau of Shenzhen Longhua District. Their unwavering support has been instrumental in facilitating a multitude of innovations, including this research, the informatics project, and the subsequent industrial application of AI techniques.\par

Gratitude to the Python community and its esteemed contributors. Your unwavering dedication to technical precision and expertise has established a robust foundation for developing cutting-edge tools for scientific computing. Your unparalleled commitment to open source collaboration and relentless innovation has fostered a thriving ecosystem that addresses unique challenges across diverse fields. Without your contributions, our progress in exploring programming practices in Python would be significantly hindered, both in terms of inspiration and the implementation of pipeline integration.\par

\clearpage

\bibliographystyle{unsrt}
\bibliography{full}

\end{document}